\documentclass[sigconf]{acmart}

\usepackage{booktabs} 

\acmConference [KDD'2018]{ACM KDD conference}{August 2018}{London, United Kingdom}
\acmYear{2018}

\begin{document}
\title{Complex Networks Classification with Convolutional Neural Netowrk}

\author{Ruyue Xin}
\affiliation{%
	\institution{School of Systems Science, Beijing Normal University}
	\streetaddress{No.19,Waida Jie,Xinjie Kou,Haiding District,Beijing}
	\city{Beijing}
	\country{China}}
\email{201621250011@mail.bnu.edu.cn}
\author{Jiang Zhang}
\affiliation{%
	\institution{School of Systems Science, Beijing Normal University}
	\streetaddress{No.19,Waida Jie,Xinjie Kou,Haiding District,Beijing}
	\city{Beijing}
	\country{China}
}
\email{zhangjiang@bnu.edu.cn}
\author{Yitong Shao} 
\affiliation{%
	\institution{School of Mathematical Sciences, Beijing Normal University}
	\streetaddress{No.19,Waida Jie,Xinjie Kou,Haiding District,Beijing}
	\city{Beijing} 
	\country{China}}
\email{201511130117@mail.bnu.edu.cn}
\renewcommand\shortauthors{Ruyue, X. et al}

\begin{abstract}
Classifying large-scale networks into several categories and distinguishing them according to their fine structures is of great importance with several applications in real life. However, most studies of complex networks focus on properties of a single network but seldom on classification, clustering, and comparison between different networks, in which the network is treated as a whole. Due to the non-Euclidean properties of the data, conventional methods can hardly be applied on networks directly. In this paper, we propose a novel framework of complex network classifier (CNC) by integrating network embedding and convolutional neural network to tackle the problem of network classification. By training the classifier on synthetic complex network data and real international trade network data, we show CNC can not only classify networks in a high accuracy and robustness, it can also extract the features of the networks automatically.
\end{abstract}

%
%



\maketitle

\section{Introduction}
Complex network is the highly simplified model of a complex system, and it has been widely used in many fields, such as sociology, economics, biology and so on. However, most of current studies focus on the properties of a single complex network\cite{Romualdo2003Statistical}, but seldom pay attention to the comparisons, classifications, and clustering different complex networks, even though these problems are also important.

Let's take the classification problem of complex networks as an example. We know that the social network behind the online community impacts the development of the community because these social ties between users can be treated as the backbones of the online community. Thereafter, we can diagnose an online community by comparing and distinguishing their connected modes. A social network classifier may help us to predict if an online community has a brilliant future or not.

As another example, let's move on to the product flows on international trade network. We know that the correct classification of products not only helps us to understand the characteristics of products, but also helps trade countries to better count the trade volume of products. But classifying and labelling each exchanged product in international trade is a tedious and difficult work. Conventional method classifies these products according to the attributes of the product manually, which is subjective. However, if a trade network classifier is built, we can classify a new product exclusively according to its network structure because previous studies point out different products have completely different structures of international trade networks.

Further, the classification problem of complex networks can be easily extended to the prediction problem. For example, we can predict the country's economic development based on a country's industrial network, or predict the company's performance based on a company's interactive structure, and so on. We can also use well-trained classifiers as feature extractors to discover features in complex networks automatically.

At present, deep learning technology has achieved state-of-art results in the processing of Euclidean Data. For example, convolutional neural network\cite{Krizhevsky2012ImageNet} (CNN) can process image data, and recurrent neural network\cite{Graves2013Speech} (RNN) can be used in natural language processing. However, deep learning technology is still under development for graph-structure data, such as social network, international trade network, protein structure data and so on.

As for complex network classification problem, there were some related researches which mainly study graph-structure data in the past. For example, kernel methods were proposed earlier to calculate the similarity between two graphs\cite{Yanardag2015Deep}. But the methods can hardly be applied on large-scale and complex networks due to the large computational complexity of these graph classification methods.

Network representation learning developed recently is an important way to study graph-structure data. Earlier works like Local Linear Embedding\cite{Roweis2000Nonlinear}, IsoMAP\cite{Joshua2000A} first constructed graphs based on feature vectors. In the past decades, some shallow models such as DeepWalk\cite{Perozzi2014DeepWalk}, node2Vec\cite{Grover2016node2vec} and LINE\cite{Tang2015LINE} were proposed which can embed nodes into high-dimensional space and they empirically perform well. However, these methods can only be applied on the tasks (classification, community detection, and link prediction) on nodes but not the whole networks.

There are also some models using deep learning techniques to deal with the network data and learn representations of networks. For example, GNN\cite{Scarselli2009The}, GGSNN\cite{Li2015Gated}, GCN\cite{Defferrard2016Convolutional} e.g. Nevertheless, these methods can also focus on the tasks on node level but not the graph level. Another shortage of GDL is the requirement of the fixed network structure background.

In this paper, we proposed a new method on complex network classification problem, which is called as a complex network classifier(CNC), by combining network embedding and convolutional neural network technique together. We first embed a network into a high-dimensional space through the DeepWalk algorithm, which preserves the local structures of the network and convert it into a 2-dimensional image. Then, we input the image into a CNN for classifying. Our model framework has the merits of the small size, small computational complexity, the scalability to different network sizes, and the automaticity of feature extraction.

The work of Antoine et al.\cite{2017arXiv170802218J} has several differences with ours. At first, our method is more simple without multi-channels and very scalable for using the classic embedding model, so that we can handle the directed and weighted networks. What's more, we apply our method more on the classification of complex network models, for that we mainly want to learn the features of classic complex network models, which is important in the development of complex network. 

The rest of the paper is organized as follows. Section 2 introduces the related research. Section 3 presents the model framework and experiments data. Section 4 shows the experiments and results and section 5 gives conclusion and discussion of the paper.

\section{Related work}

\subsection{Complex network}
Complex network focuses on the structure of individuals' interrelation in system and is a way to understand the nature and function of complex system. Studies of complex networks started from regular networks, such as Euclidean grid or nearest neighbor network in the two-dimensional plane. In 1950, Erdos and Renyi proposed random network theory. In 1998, Watts\cite{Watts1998Collectivedynamics} and Barab\'asi\cite{Barab1999Emergence} proposed small-world and scale-free network models, respectively, which depict real life networks better. Researchers have summarized the classic complex network model includes regular networks, random networks, small-world networks, scale-free networks, and proposed the properties of networks such as average path length, aggregation coefficient and degree distribution. Recent studies mainly focus on network reconstruction, network synchronization etc., and few studies focus on the classification of complex networks.

\subsection{Network classification}
Classification of network data has important applications such as protein-protein interaction, predicting the functionality of the chemical compounds, diagnosing communities and classifying product trading networks. In the network classification problem, we are given a set of networks with labels, and the goal is to predict the label of a new set of unlabeled networks. The kernel methods developed in previous research are based on the comparison of two networks and similarity calculation. The most common graph kernels are random walk kernels\cite{Kashima2003Marginalized}, shortest-path kernels\cite{Borgwardt2005Shortest}, graphlet kernels\cite{Shervashidze2009Efficient}, and Weisfeiler-Lehman graph Kernels\cite{Shervashidze2011Weisfeiler}. However, the main problem of graph kernels is that they can not be used in large-scale and complex networks for the expensive calculation complexity.

\subsection{Deep learning on graph-structure data}
CNN is the most successful model in the field of image processing. It has achieved good results in image classification\cite{Krizhevsky2012ImageNet}, recognition\cite{Simonyan2014Very}, semantic segmentation\cite{Shelhamer2017Fully} and machine translation\cite{Kalchbrenner2014A} and can independently learn and extract features of images. 

However, it can only be applied on regular data such as images for fixed size. As for graph-structure data, researchers are still trying to solve it with deep learning methods recently. For example, in order to apply the convolutional operation on graphs, \cite{Bruna2013Spectral} proposed to perform the convolution operation on the Fourier domain by computing the graph decomposition of the Laplacian matrix. Furthermore, \cite{Henaff2015Deep} introduces a parameterization of the spectral filters. \cite{Defferrard2016Convolutional} proposed an approximation of the spectral filter by Chebyshev expansion of the graph Laplacian. \cite{Kipf2016Semi} simplified the previous method by restricting the filters to operate in a 1-step neighborhood around each node. 

However, in all of the aforementioned spectral approaches, the learned filters based on the laplacian eigenbasis is dependent on the graph structure. Thus, a model trained on a specific structure can not be directly applied to a graph with a different structure. We know that a complex network classification problem often includes many samples and each sample has one specific network structure, so we can not directly use GCN to classify networks.

\subsection{Network representation learning}
Representation learning has been an important topic in machine learning for a long time and many works aim at learning representations for samples. Recent advances in deep neural networks have witnessed that they have powerful representation abilities and can generate very useful representations for many types of data. 

Network representation learning is an important way to preserve structure and extract features of network through network embedding, which maps nodes into a high-dimensional vector space based on graph structure. And the vector representations of network nodes can be used for classification and clustering tasks.

There are some shallow models proposed earlier for network representation learning. DeepWalk \cite{Perozzi2014DeepWalk} combined random walk and skip-gram to learn network representations. LINE\\ \cite{Tang2015LINE} designed two loss functions attempting to capture the local and global network structure respectively. Node2Vec\cite{Grover2016node2vec} improved DeepWalk and proposed a 2-order random walk to balance the DFS and BFS search.

The most important contribution of network representation learning is that it can extract network features which provide a way to process network data. So we consider to use the features extracted by the embedding methods to solve the network classification problem. We recognize DeepWalk is a classic and simple model which can represent network structure and has high efficiency when dealing with large-scale networks. Besides, the Random Walk process in DeepWalk which obtains the sequences of networks is adaptable to different networks, for example, we can easily change the random walk mechanism for international trade network which is directed and weighted. So we combine the networking representation learning and deep learning method to develop our model, which can perform well in the complex network classification task.

\section{Methods of network classification}
\subsection{the model}
Our strategy to classify complex networks is to convert networks into images, and use the standard CNN model to perform the network classification task. Due to the development of network representation techniques, there are a bunch of algorithms to embed the network into a high dimensional Euclidean space. We select DeepWalk algorithm \cite{Perozzi2014DeepWalk}, which is proposed by Bryan Perozzi et al to obtain the network representation. The algorithm will generate numeric node sequences by performing large-scale random walks on the network. After that, the sequences are fed into the SkipGram + Negative Sampling algorithm to obtain the Euclidean coordinate representation of each node.

Obviously high-dimensional space representation is hard to be processed, thus we use the PCA algorithm to reduce the dimension of node representations into 2-dimensional space. However, the set of nodes is a point cloud which is still irregular and cannot be processed by CNN, thus we rasterize the 2-dimensional representation into an image. We divide all the areas covered by the 2-dimensional scatter plot into a square area with $48 * 48$ grids and then count the number of nodes in each grid as the pixel grayscale. After that, a standard gray scale image is obtained. The reason why we do not embed the network into 2-dimensional space directly is because we believe that doing so may lose less information, particularly for the local structures. This method can also be applied on directed and weighted networks like international trade flow networks. By adjusting the probabilities according to the weight and direction of each edge for a random walk on a network, we can obtain an embedded image.

\begin{figure}[!h]
	\includegraphics[width=\columnwidth]{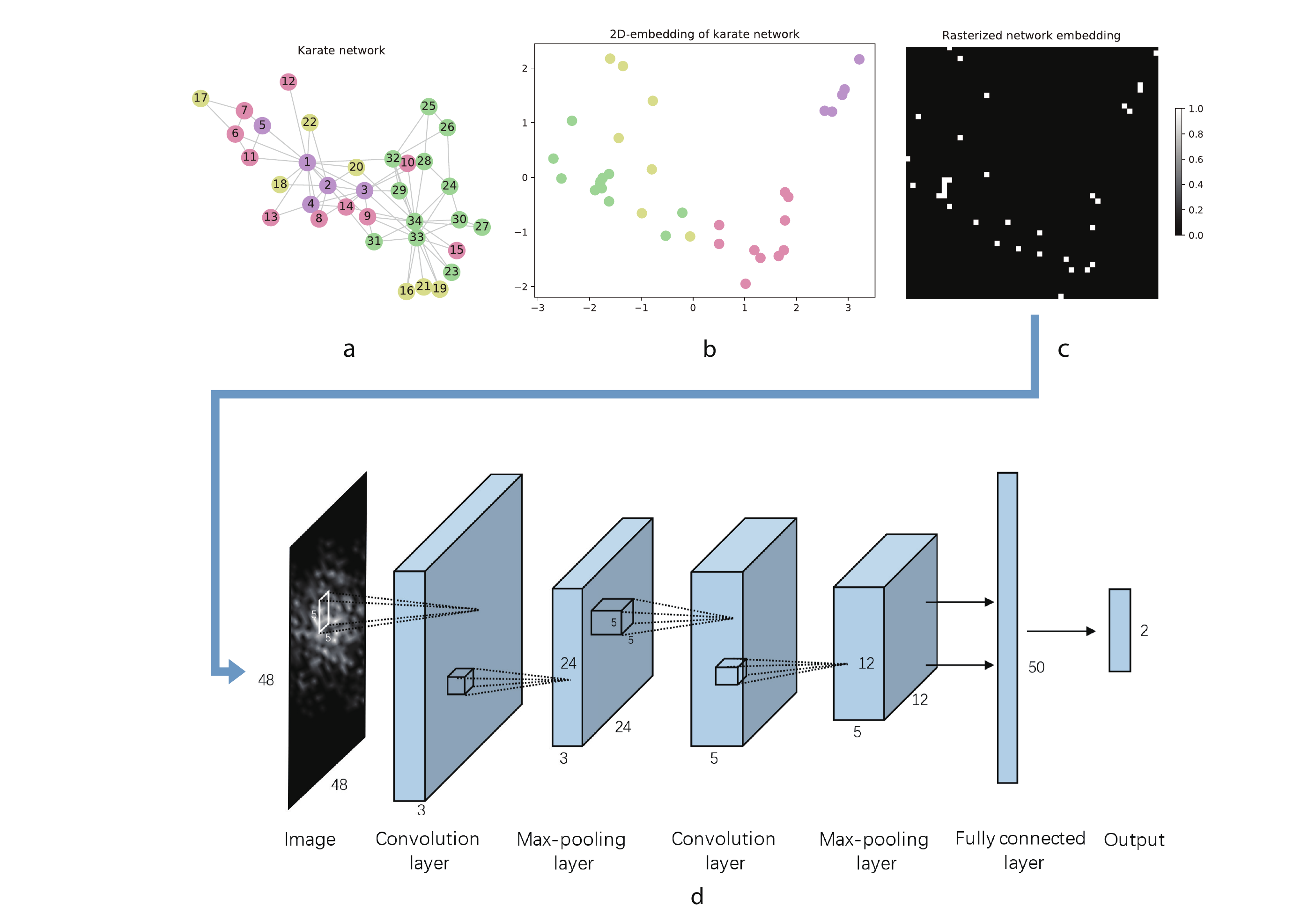}
	\caption{The pipeline of CNC algorithm. (a) The original input network. (b) The embedding of the network with DeepWalk algorithm. In DeepWalk algorithm, to obtain enough "corpus", we set the number of walks to 10000 times and the sequence length to 10, and embed the network into the 20-dimensional space and then reduce it to 2-dimensional space. (c) The rasterized image from the 2D-embedding representation of the netowrk. (d) The CNN architecture of CNC algorithm, which includes one input image, two convolutional-pooling layers, one fully-connected layer and one output layer. The sizes of the convolutional filters are $5 * 5$, of the pooling operation is $2 * 2$. The first layer has 3 convolutional filters, and the second layer has 5 convolutional filters, and the fully connected layer has 50 units. In all complex networks classification experiments, we set the learning rate = 0.01 and mini-batch = 100. The CNN architecture is selected as mentioned to minimize the computational complexity as well as keeping the classification accuracy.
	}
	\label{fig:1}
\end{figure}

The final step is to feed the representative images into a CNN classifier to complete the classification task. Our convolutional neural network architecture includes two convolutional layers (one convolutional operation and one max-pooling operation) and one fully-connected layer and one output layer. The whole architecture of our model can be seen in Fig.\ref{fig:1}.

\subsection{Experiment data}

A large number of experimental data is needed to train and test the classifier, thus we use both synthetic networks generated by network models and empirical networks to test our model.

\subsubsection{Synthetic data}

The synthetic networks are generated by well known BA and WS models. According to the evolutionary mechanism of BA model, which iteratively adds $m = 4$ nodes and edges at each time, and the added nodes will preferentially link to the existing nodes with higher degrees until $n = 1000$ nodes are generated, and the average degree $<E>$ of the generated network is about 8 which is close to the degree of real networks\cite{Wang2003Complex}. We then use WS model ($n = 1000$, the number of neighbors of each node $k = 8$, and the probability of reconnecting edges $p = 0.1$) to generate a large amount of small-world networks with the same average degrees as in BA model.

We then mix the generated 5600 BA networks and WS networks, respectively. And we separate the set of networks into training set (with 8000 networks), validation set (with 2000 networks), and test set (with 1200 networks).

\subsubsection{Empirical data}

Product specific international trade networks are adopted as the empirical data to test our classifier, the dataset is provided by the National Bureau of Economic Research (\url{http://cid.econ.ucdavis.edu/nberus.html}) and covers the trade volume between countries of more than 800 different kinds of products which are all encoded by SITC4 digits from 1962 to 2000. Notice that the international trade network is a weighted directed network, in which the weighted directed edges represent the volumes of trading flows between two countries. Thus, the random walk in DeepWalk algorithm should be based on the weights and directions of edges. We train the CNC to distinguish the food products and chemicals products. Each product class contains about 10000 networks obtained by the products and the products combinations within the category.

\section{Experiments and Results}

We conduct a large number network classification experiments, and the results are present in this section. On the synthetic networks, we not only show the classification results, but also present how the CNC can extract the features of networks, and the robustness of the classifier on network sizes. On the empirical networks, we show the results that our CNC apply on the trade flow networks which are directed weighted networks. 

\begin{figure}[H]
	\centering
	\begin{minipage}[b]{.45\linewidth}
		\centering
		\includegraphics[width=\linewidth]{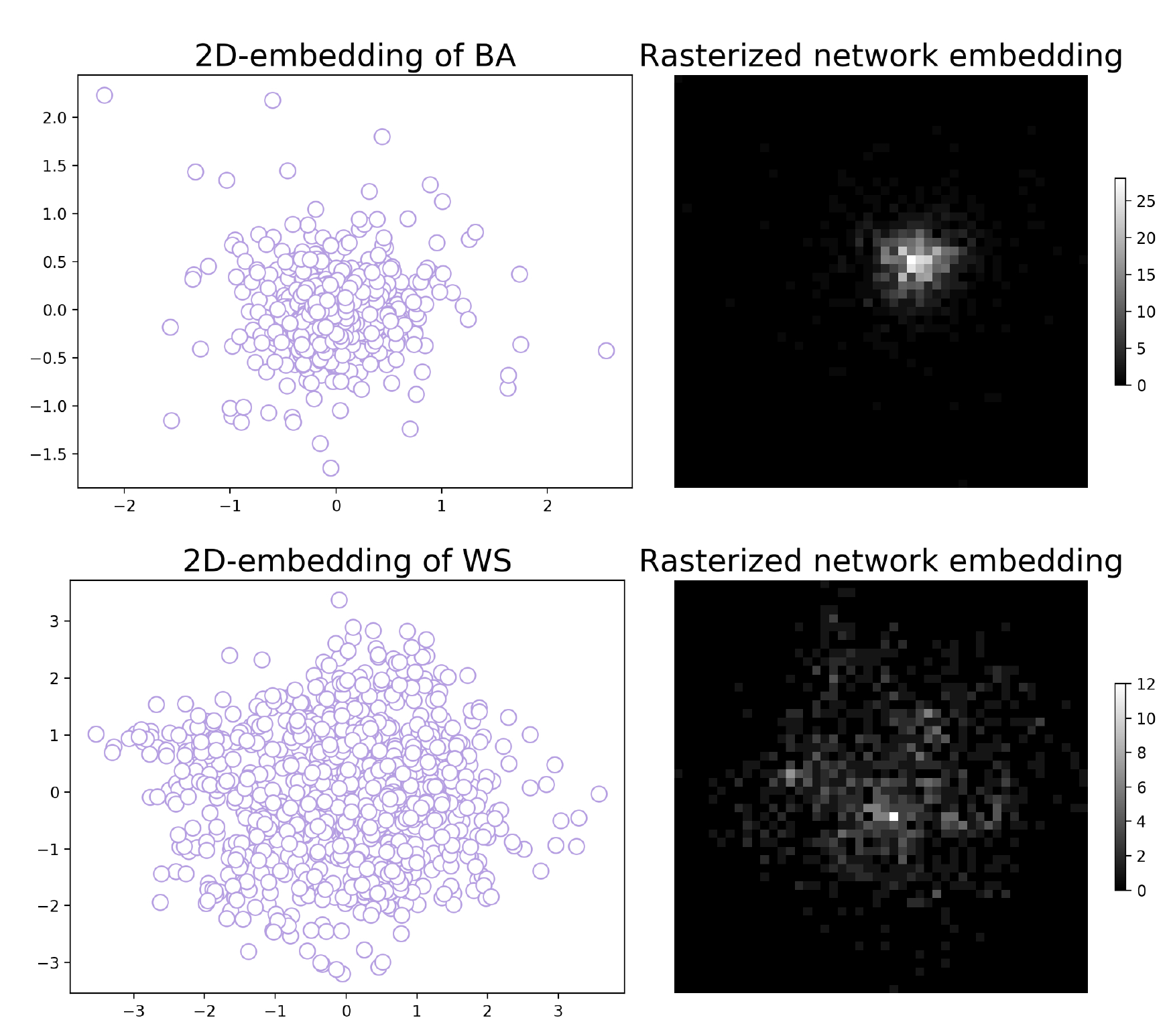}\\
		a\\
		\vspace{0.2cm}
		\includegraphics[width=.9\linewidth]{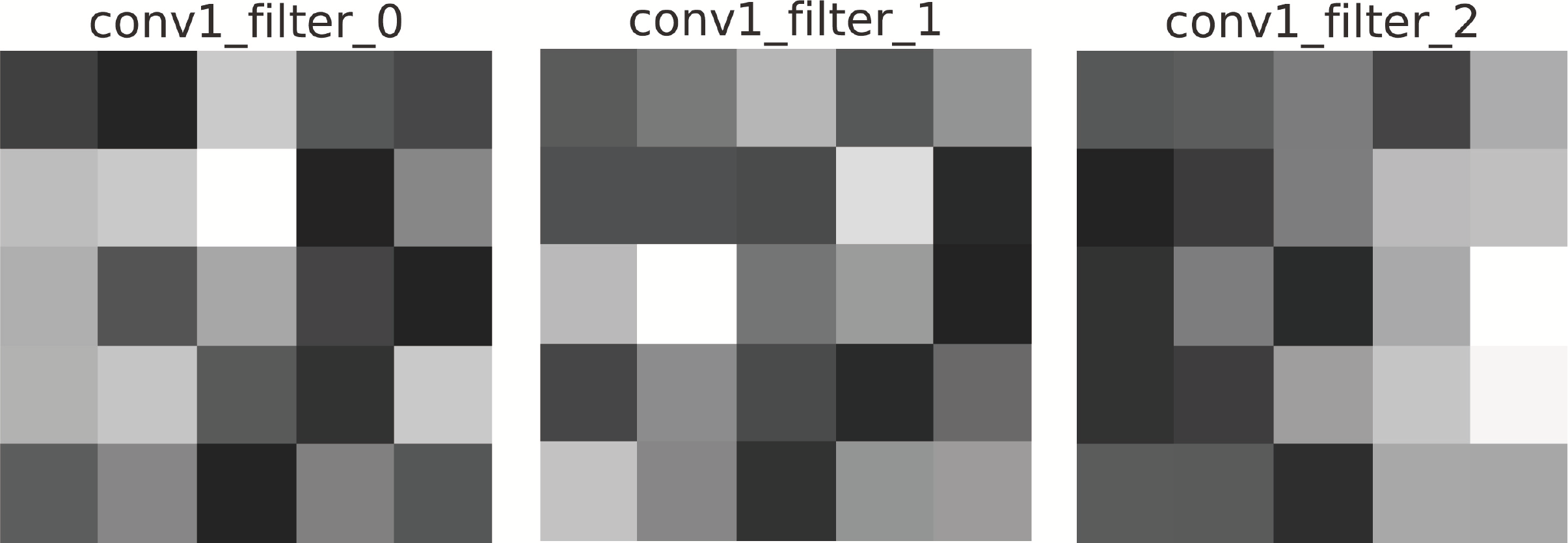}\\
		b\\
	\end{minipage}
	\quad
	\begin{minipage}[b]{.45\linewidth}
		\centering
		\includegraphics[width=.9\linewidth]{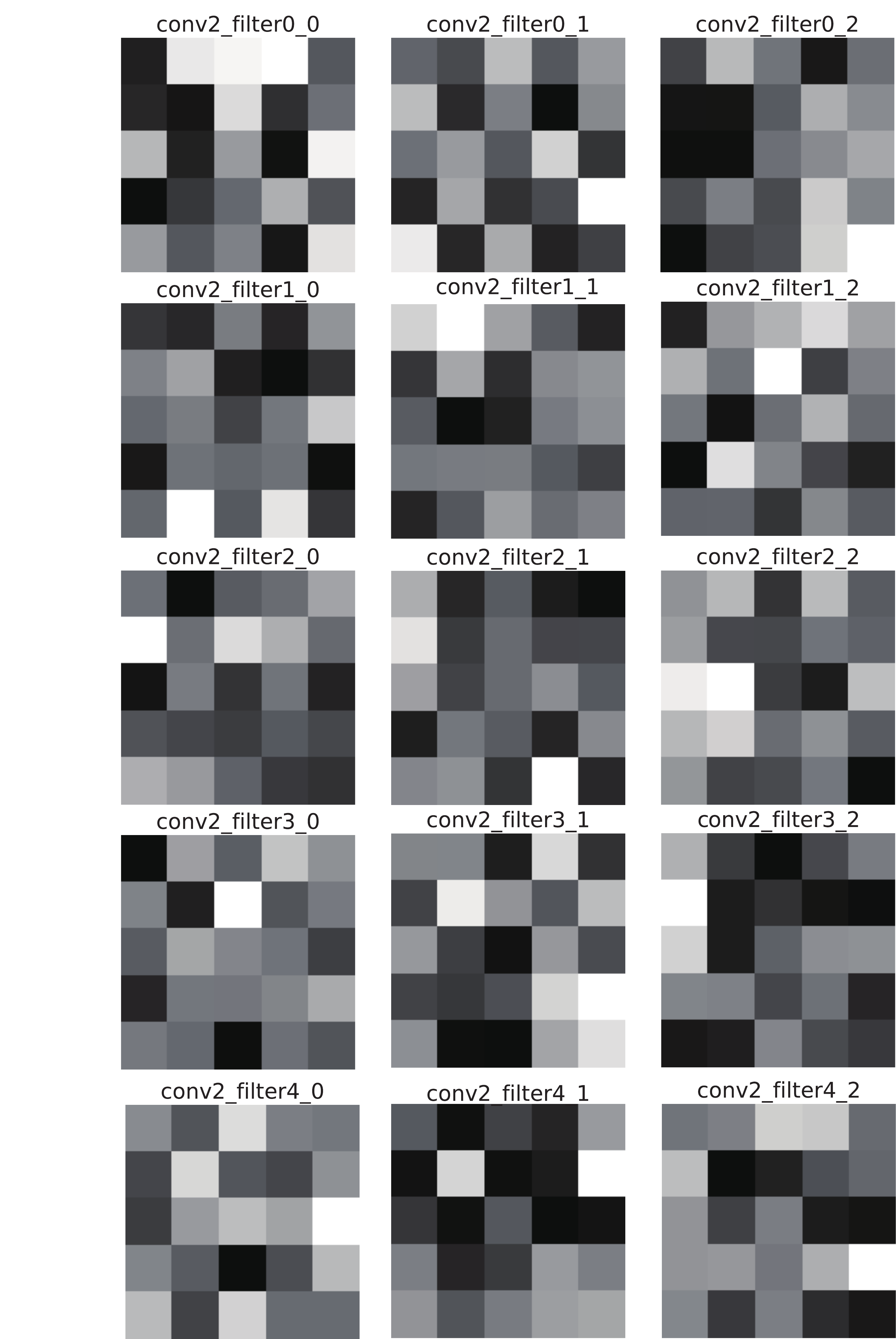}\\
		c\\
	\end{minipage}
	\caption{\label{fig:2}(a) The 2D representations and rasterized images of a BA network (upper) and a WS network (bottom). (b) Visualization of the three filters of the first convolutional layer. (c) Visualization of the five filters (size of $5*5*3$) of the second convolutional layer.}
\end{figure}

\subsection{Classification experiments on synthetic networks}
\subsubsection{BA and WS classification experiments}
The first task is to apply CNC to distinguish BA network and WS network. Although we know the BA network is a scale-free network, and WS network is a small-would network with high clustering coefficient, machine does not know. Thus this series experiments show the possibility that the CNC network can extract the key features to distinguish the two kinds of networks. We generate 5600 BA networks with $n=1000$, $m=4$ and 5600 WS networks with the same size ($n=1000$, $k=8$) and $p=0.1$, respectively. And we mix these networks to form the data set which is further randomly separated into training set (with 8000 networks), validation set (with 2000 networks), and test set (with 1200 networks).
Figure \ref{fig:6}(a) shows the decay of the loss on training set and error rate of validation set. Finally, we obtain the average error rate 0.1\% on the test set. So we can say the model can distinguish the BA network and the WS network accurately.

To understand what has been learnt by our CNC model, we can visualize the feature maps extracted from the network representations by the filters of the CNN, which are visualized in Fig.\ref{fig:2}. However, it is hard to read meaningful information because the network structure cannot be corresponded to the images.

\begin{figure}[H]
	\includegraphics[width=\columnwidth]{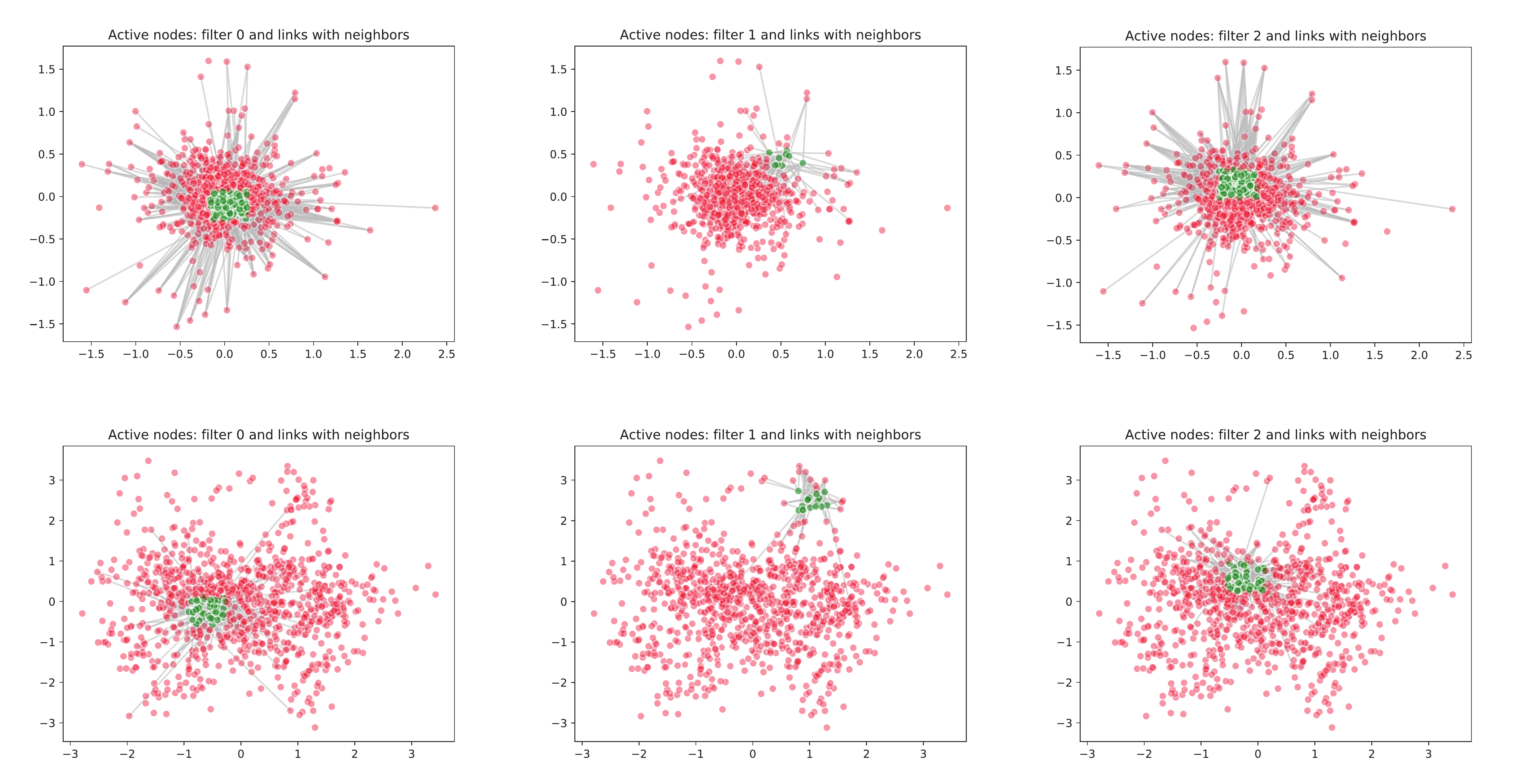}
	\caption{We show the active nodes corresponding to the highlighted areas in the feature maps of the 3 filters of the first convolutional layer when inputting a typical BA network and WS network respectively. We draw the activated nodes (the green points) and their links with other nodes as the background for the two networks. (Upper: scale-free network. Bottom: small-world network).}
	\label{fig:3}
\end{figure}

To understand what the filters do, we need combine the network structure and the feature map. Therefore, we try to map the highlighted areas in feature maps of each filter on the nodes sets of the network. That is, we wonder which parts of the networks and what kind of local structures are activated by the first convolutional layer filters. We compare the activation modes for the two model networks as input, and the results are shown in Fig.\ref{fig:3}. By observing and comparing these figures, we find that the convolutional filters of the first layer has learnt to extract the features of the network in different parts. As shown in Fig.\ref{fig:3}, Filter 0 is extracting the local clusters with medium density of nodes and connections; and Filter 1 tries to extract the local clusters with sparse connections; while Filter 2 tries to extract the local clusters with dense nodes and connections.

By comparing BA and WS model networks, we can observe that the locations and the patterns of the highlighted areas are different. The local areas with dense nodes and connections (Filter 0) locate the central area of the network representation for both BA network and WS network. The local structures with sparse nodes and connections locate the peripheral area which is close to the edges of the image for the WS network, but it is in the central area for the BA network. This combination of the activation modes on feature maps can help the higher level filters and fully connected layer to distinguish the two kinds of networks.

\subsubsection{Small world networks classification}

One may think to distinguish the BA and WS networks is trivial because they are two different models at all. Our second experiment will consider whether the classifier can distinguish networks generated by different parameters of the same model, which is harder than the previous task.

In order to verify the discriminant ability of the model on this task, we use the WS model to generate a large number of experimental networks by changing the value of edge reconnection probability p from 0 to 1 in a step of 0.1, and then we mix the networks with two discriminant $p$ values, eg. $p=0.1$ and $p=0.6$, and we train the CNC for networks, and test their discriminant ability on the test sets. 

\begin{figure}[H]
	\includegraphics[width=\linewidth]{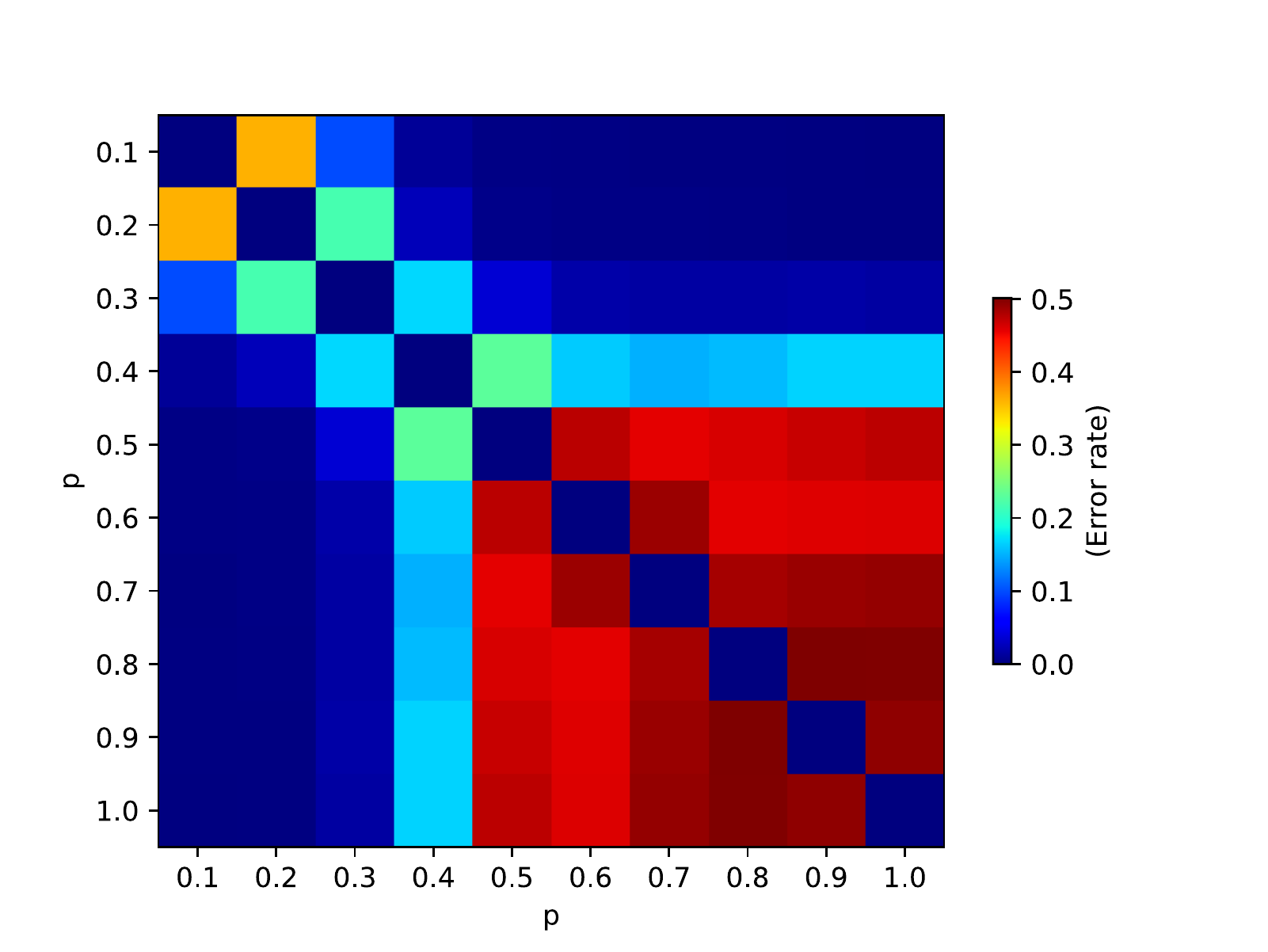}
	\caption{the classification results of each two small-world networks with different p value.}
	\label{fig:4}
\end{figure}

We systematically do this experiment for any combination of the networks with each two probabilities, and the results are shown in Fig.\ref{fig:4}. We can see that the networks generated by $p$ values less than 0.3 and p values greater than or equal to 0.4 are easier to be distinguished. Interestingly, there is a sudden change for the error rate at $p=0.4$. For the two networks with $p>0.4$, the classifier cannot distinguish them. The reason behind this phenomenon may be due to the phase transition of the link percolation in random networks because the WS networks with $p>0.5$ may be treated as random networks.

\begin{figure}[H]
	\includegraphics[width=\columnwidth]{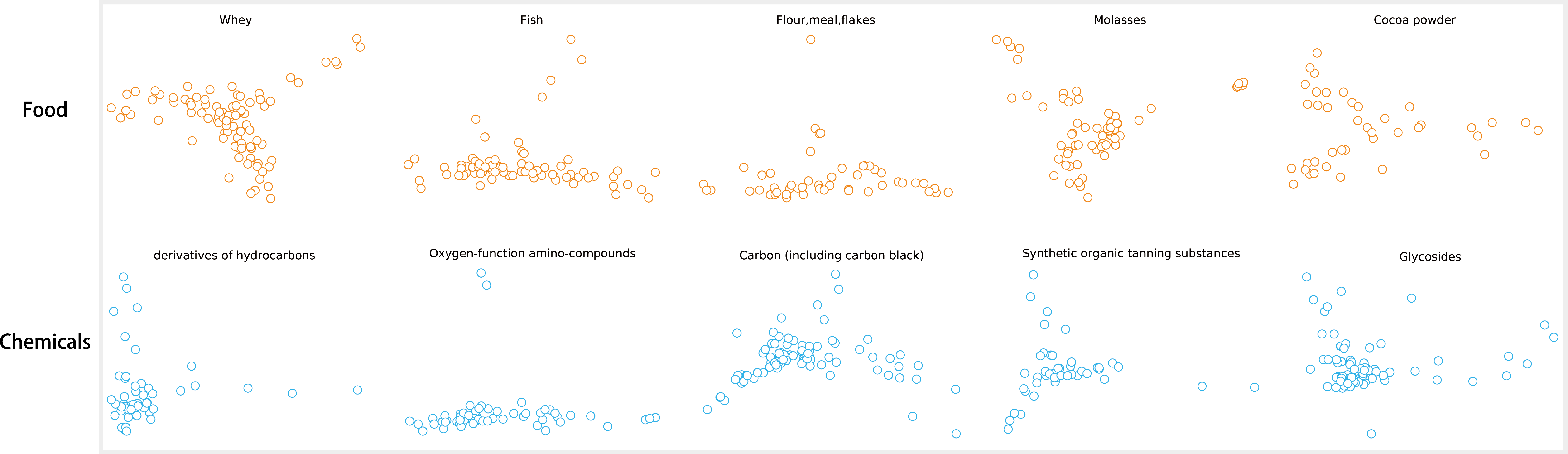}
	\caption{Network representations of 10 selected products in two classes: food (upper) and chemicals (bottom).}
	\label{fig:5}
\end{figure}

\subsection{Classification on trade flow networks}

\begin{figure}[H]
	\begin{minipage}{0.48\columnwidth}
		\centerline{\includegraphics[width=4.0cm]{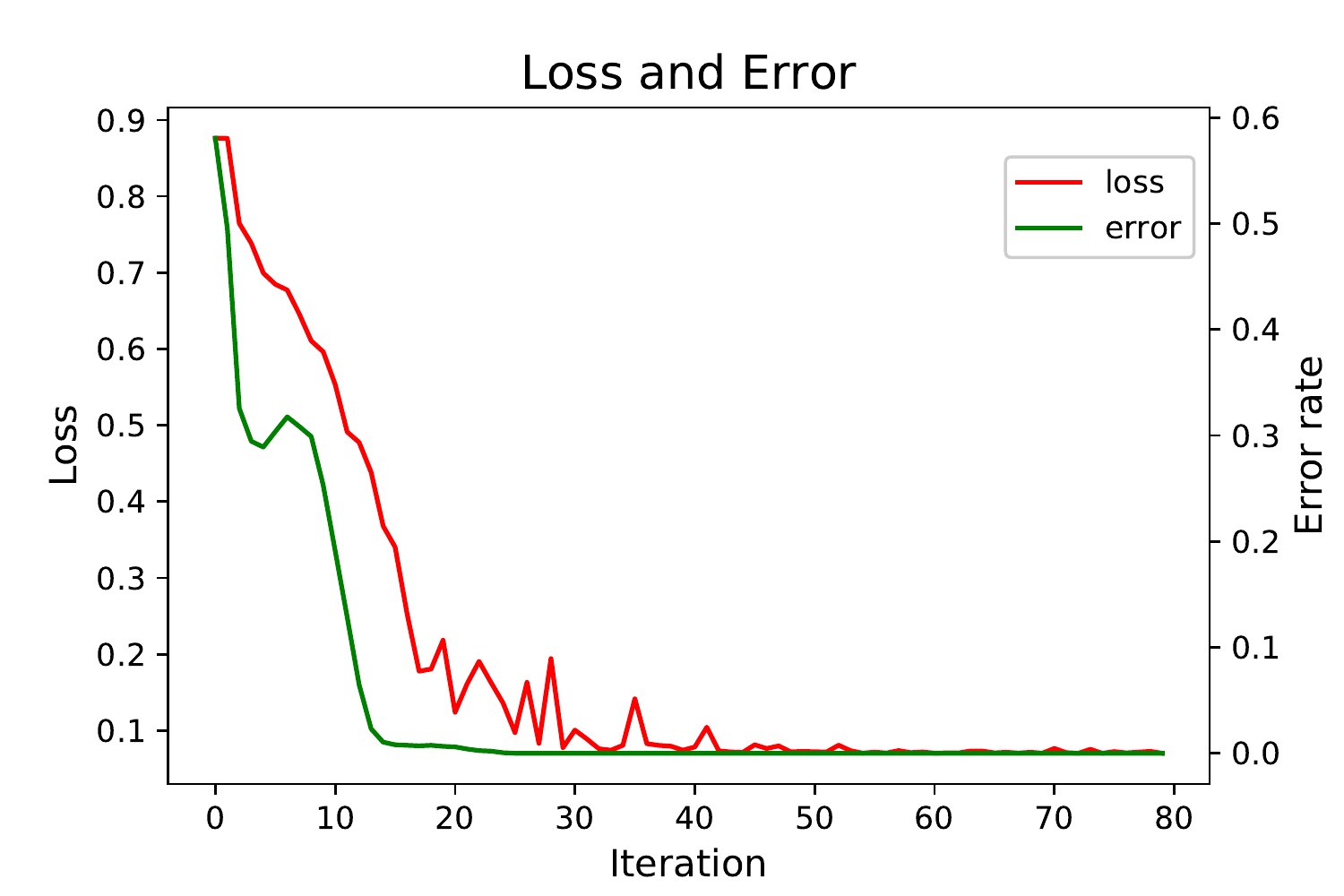}}  
		\centerline{a} 
	\end{minipage}
	\hfill
	\begin{minipage}{0.48\columnwidth}
		\centerline{\includegraphics[width=4.0cm]{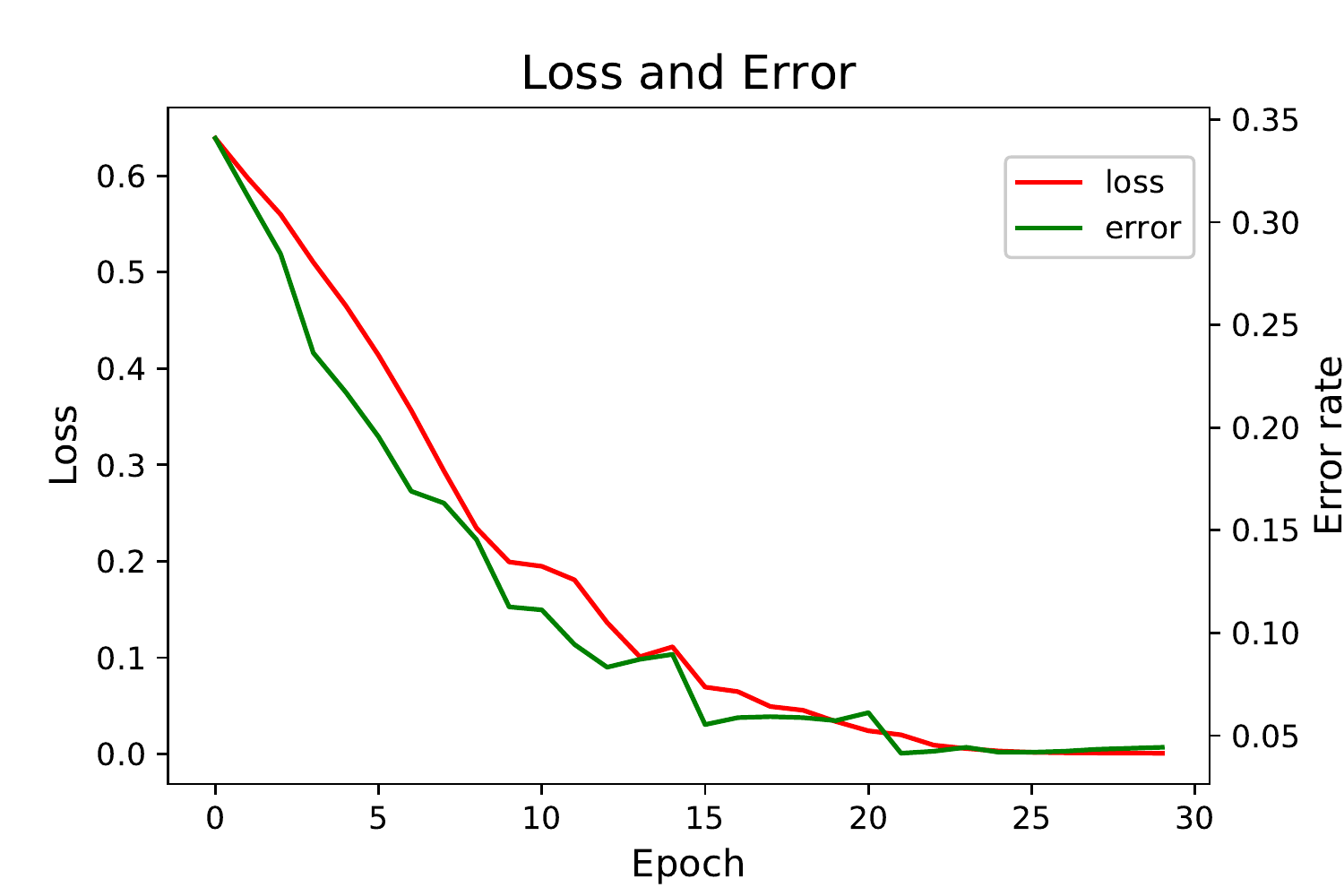}}  
		\centerline{b} 
	\end{minipage}
	\caption{Plots of loss and validation error rate of the classification task on BA v.s. WS models (a) and the classification task for food v.s. chemicals products (b).}
	\label{fig:6}
\end{figure}

We want to verify the effectiveness of the model on empirical networks. We conduct a classification on international trade flow networks with the dataset obtained from the National Bureau of Economic Research (\url{http://cid.econ.ucdavis.edu/nberus.html}). This data covers the trade volume and direction information between countries of more than 800 different kinds of products which are all encoded by SITC4 digits from 1962 to 2000. We select food and chemicals products as two labels for this experiment, and their SITC4 encoding starts with 0 and 5 respectively. For example, 0371 is for prepared or preserved fish and 5146 is for oxygen-function amino-compounds. Fig.\ref{fig:5} shows the 2-dimensional representation of the 10 products for two categories. After pre-processing, the number of the food trade networks is 10705 (including products and product combinations with SITC4 digits starting with 0) and the chemicals trade network is 10016 (including products and product combinations with SITC4 digits starting with 5). Then, we divide them into training set, validation set and test set according to the ratio of 9: 1: 1. During the training, we adjust the network parameters to 15 convolutional filters in the first layer and 30 convolutional filters in the second layer, 300 units of the full-connect layer. Fig.\ref{fig:6}(b) shows that the classification error rate can be cut down to 5\%.

\begin{figure}[H]
	\begin{minipage}{0.48\columnwidth}
		\centerline{\includegraphics[width=4.0cm]{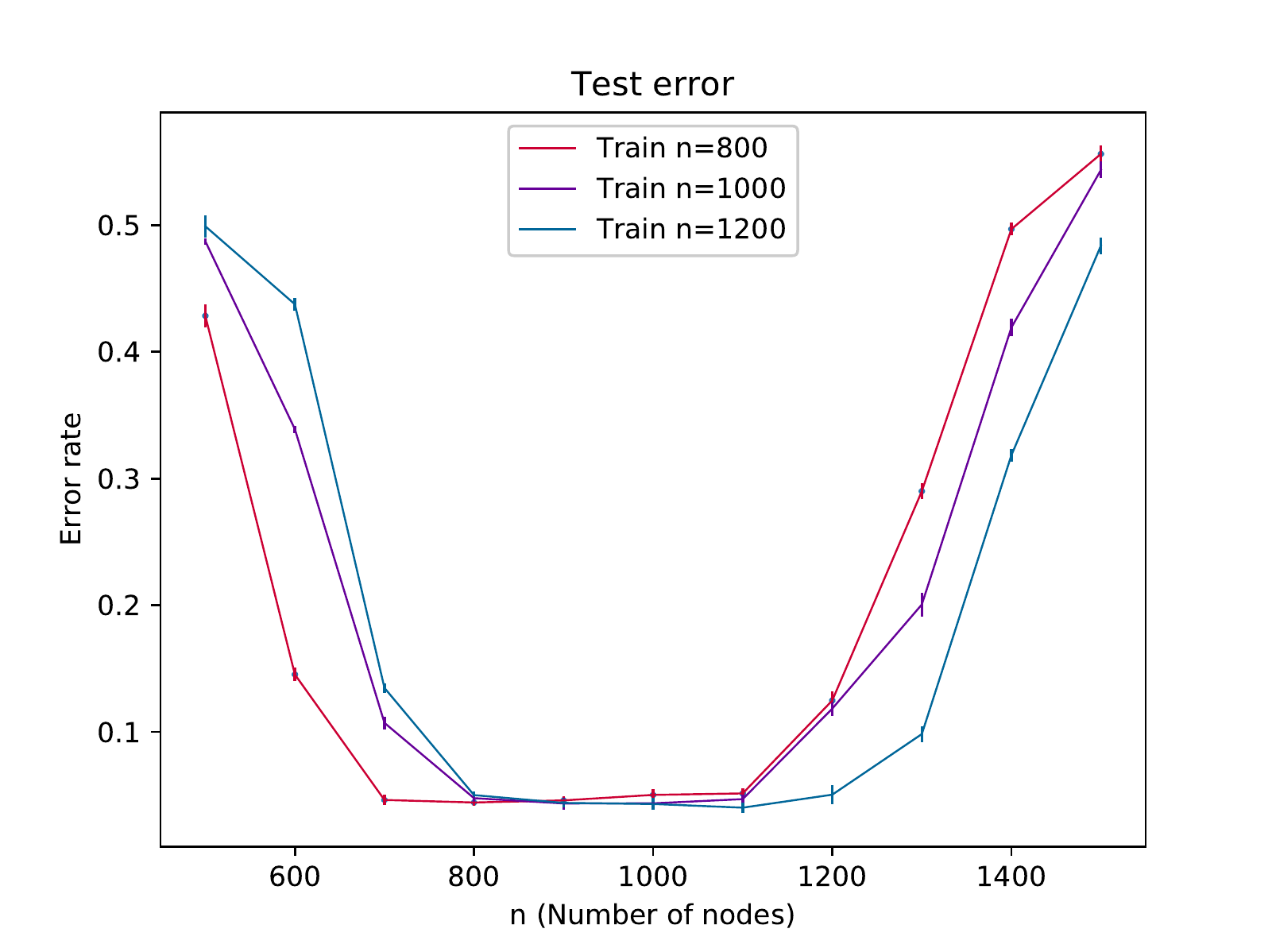}}  
		\centerline{a} 
	\end{minipage}
	\hfill
	\begin{minipage}{0.48\columnwidth}
		\centerline{\includegraphics[width=4.0cm]{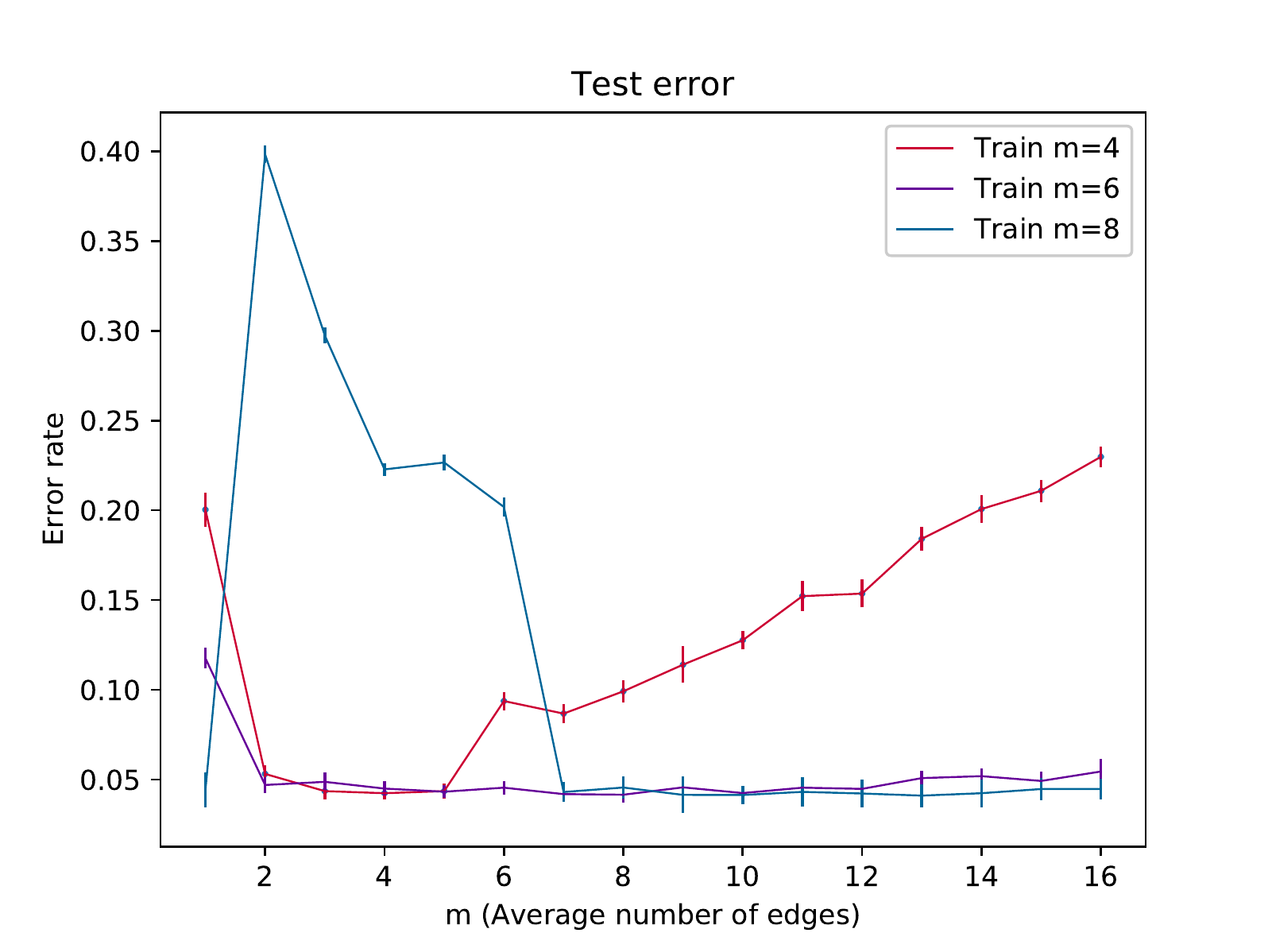}}  
		\centerline{b} 
	\end{minipage}
	\caption{The dependence of the error rates on the number of nodes (left) and the number of edges(right) in the robustness experiments. (a) In test set, we set $n$ (number of nodes) = $[500,600,700,\cdots,1500]$, and we also retrain $n=800$ and $n=1200$ and test them with the different $n$ test set. (b) In test set, we set $m$ (average number of edges) = $[1,2,3,\cdots,16]$, and we also retrain $m=6$ and $m=8$ and test them with the different $m$ test set. }
	\label{fig:7}
\end{figure}

\subsection{Robustness on sizes of the network}

Our model has good classification performances on both synthetic and empirical data. Next, we want to test the robustness of the classification on different sizes (numbers of nodes and edges). Note that all the experiments performed in classification experiments contain the model networks with identical numbers of nodes and edges. Nonetheless, a good classifier should extract the features which are independent on size. Therefore, we examine the robustness of the classifier on various network sizes which are different from the training sets. In these experiments, we first apply the trained classifier for BA and WS networks with $n=1000$ nodes and average degree $<E>=8$ , on new networks different numbers of nodes and edges. We generate 600 mixed networks by BA and WS models with parameters $m$ from $[1, 2, 3,\cdots,16]$ for the BA model and $k$ from $[2, 4, 6,\cdots,32]$ for the WS model as test set such that their average degrees are similar. 

We systematically compare how the number of nodes (left) and edges (right) on the test sets influence the error rates as shown in Fig.\ref{fig:7}. At first, we observe that the error rates are almost independent on small fluctuations of the number of nodes. However, the error rates increase as larger size differences are in the test data. This manifest our classifiers are robust on the size of the networks.

Nevertheless, there are sudden changes for the variants on the number of edges, which indicates that the number of edges has larger impacts on the network structure.  We observe that there is a sudden drop on error rates with increase of $m$ for the test set when $m=8$ for the training set. Through observing the network embedding grow we know that he reason behind this sudden change is the emergence of the multi-center on the representation space for the BA model. Therefore, the number of links can change the overall structure in the scale free network, and this makes our classifier working worse. Another interesting phenomenon is the error rates can keep small when the number of edges increase when $m$ in the training set is set to 8. Therefore, the classifiers training on the dense networks are more robust on the variance on edge densities.

\section{Conlusion and Discussion}

In this paper, we propose a model, which mainly incorporates DeepWalk and CNN, to solve the network classification problem. With DeepWalk, we obtain an image for each network, and then we use CNN to complete the classification task. Our method is independent on the number of network samples, which is a big limitation for the kernel methods on graph classification. We validate our model by experiments with the synthetic data and the empirical data, which show that our model performs well in classification tasks. In order to further understand the network features extracted by our model, we visualize the filters in CNN and we can see that CNN can capture the differences between WS and BA networks. Furthermore, we test the robustness of our model by setting different sizes for traineding and testing. The biggest advantage of our model is that our model can deal with networks with different structures and sizes. In addition, the architecture of our model is small and the computational complexity is low.

There are several potential improvements and extensions to our model that could be addressed as future works. For example, we can develop more methods to deal with the network features in high-dimensional space. Besides, we think that our model can be applied to more classification and forecasting tasks in various fields. Finally, we believe that extending our model to more graph-structure data would allow us to tackle a larger variety of problems.

\begin{acks}
	
  The authors would like to thank the referees for
  their valuable comments and helpful suggestions. The work is
  supported by the \grantsponsor{GS501100001809}{National Natural
    Science Foundation of
    China}{http://dx.doi.org/10.13039/501100001809} under Grant
  No.:~\grantnum{GS501100001809}{61673070}
  and Beijing Normal University Interdisciplinary Project.

\end{acks}

\bibliographystyle{ACM-Reference-Format}
\bibliography{cite}

\end{document}